\documentclass[conference]{IEEEtran}
\IEEEoverridecommandlockouts
\usepackage{cite}
\usepackage{amsmath,amssymb,amsfonts}
\usepackage{algorithm}
\usepackage{algorithmic}
\usepackage{graphicx}
\usepackage{textcomp}
\usepackage{xcolor}
\usepackage{multirow}
\usepackage{latexsym}
\usepackage{color}
\newcommand{\red}[1]{\textcolor{red}{#1}}
\usepackage{array}
\usepackage{footnote}
\makesavenoteenv{tabular}
\usepackage{CJKutf8}
\graphicspath{{Images/}}

\makeatletter

\newcommand{\Rmnum}[1]{\expandafter\@slowromancap\romannumeral #1@}
\makeatother
\def\BibTeX{{\rm B\kern-.05em{\sc i\kern-.025em b}\kern-.08em
    T\kern-.1667em\lower.7ex\hbox{E}\kern-.125emX}}
\begin{document}
\begin{CJK}{UTF8}{gbsn}
\bibliographystyle{unsrt}
\title{Emergent Incident Response for Unmanned Warehouses with Multi-agent Systems*\\
{\footnotesize \textsuperscript{*}Note: Sub-titles are not captured in Xplore and
should not be used}
\thanks{Identify applicable funding agency here. If none, delete this.}
}

\author{\IEEEauthorblockN{1\textsuperscript{st} Yibo Guo}
\IEEEauthorblockA{\textit{School of Computer} \\
\textit{and Artificial Intelligence} \\ 
\textit{Zhengzhou University}\\
Zhengzhou, China }
\and
\IEEEauthorblockN{2\textsuperscript{nd} Mingxin Li}
\IEEEauthorblockA{\textit{School of Computer} \\
\textit{and Artificial Intelligence} \\
\textit{Zhengzhou University}\\
Zhengzhou, China\\
}
\and
\IEEEauthorblockN{3\textsuperscript{rd} Jingting Zong}
\IEEEauthorblockA{\textit{School of Computer} \\
\textit{and Artificial Intelligence} \\
\textit{Zhengzhou University}\\
Zhengzhou, China\\
}
\and
\IEEEauthorblockN{4\textsuperscript{th} Mingliang Xu$^{\ast}$}
\IEEEauthorblockA{\textit{School of Computer} \\
\textit{and Artificial Intelligence} \\
\textit{Zhengzhou University}\\
Zhengzhou, China \\
}
}
\maketitle

\begin{abstract}
Unmanned warehouses are an important part of logistics, and improving their operational efficiency can effectively enhance  service efficiency. However, due to the complexity of unmanned warehouse systems and their susceptibility to errors,   incidents may occur during their operation, most often in inbound and outbound operations, which can  decrease   operational efficiency. Hence it is crucial to  to improve  the response to such  incidents. This paper proposes a collaborative optimization algorithm for emergent incident response based on Safe-MADDPG. To meet   safety requirements during emergent incident response, we investigated the intrinsic hidden relationships between various factors. By obtaining constraint information of agents during the emergent incident response process and   of the dynamic environment of unmanned warehouses on agents, the algorithm reduces   safety risks   and avoids the occurrence of chain accidents; this enables an unmanned system to complete emergent incident response tasks and achieve its optimization objectives: (1) minimizing the losses caused by emergent incidents; and (2) maximizing the operational efficiency of inbound and outbound operations during the response process. A series of experiments   conducted in a simulated unmanned warehouse scenario   demonstrate the effectiveness of the proposed method.
\end{abstract}


\section{Introduction}
With economic   development and   innovations in the internet, big data, blockchain, and artificial intelligence, e-commerce has rapidly developed. This has resulted in a rapid increase in logistics tasks, with associated problems such as the improvement of logistical efficiency,   error rates, and   labor costs, such as through  unmanned warehouses and   an increased proportion of unmanned and intelligent logistics. However,  high-density tasks and high-intensity operations may cause   incidents such as equipment failures and shelf collapses. The inbound and outbound operation link is a common unexpected event in the operation of unmanned warehouses, which can decrease  efficiency.

Emergent incidents are events that may affect the efficiency of inbound and outbound operations in unmanned warehouses and require immediate attention. They can be categorized according to whether they  cause damage to the surrounding environment. Those  not likely to cause damage   include the following:

(1) Goods falling or getting stuck during automated guided vehicle (AGV) transportation, causing the AGV to stop running; 

(2) Deviation of a shelf position or unbalanced placement of goods, such that an AGV is unable to lift a shelf, or   its driving trajectory   deviates; 

(3) AGV protection failures, such as against collisions, collision avoidance scheduling, or communication interruption. 

Emergent incidents that are likely to cause damage  will spread  after they occur, causing   incidents in multiple areas, including: 

(1) Fire in automated equipment;

(2) Explosion of flammable and explosive goods.

We choose fire in automated equipment as an example of the second category, whose response   is  carried out by professionals who arrive on-site for targeted disposal, which may affect   operational efficiency. In a warehouse  with equipment   running at high speeds, the use of unmanned systems in response to incidents has a relatively small impact on   efficiency. Therefore, the   response in   this paper uses agents composed of unmanned systems. The   task of emergent incident response is to   quickly control the source of danger, whose goals are to minimize   losses and  achieve the highest operational efficiency of inbound and outbound operations.
There has been extensive research on emergent incident scheduling problems in complex environments,  showing that the early resolution of   incidents can avert serious disasters \cite{1-1}. \cite{1-2} studied the coordination and control of dynamic abnormal emergent incident detection in multi-robot systems. A distributed multi-agent dynamic abnormal detection and tracking method was proposed,   integrating full map coverage and multiple dynamic abnormal tracking with a set of rescue robots. During the search phase, robots are guided to locations with higher probabilities of emergent incidents. Reinforcement learning (RL) has been applied to dynamic scheduling problems in complex environments; however, some studies have shown that its learning process may lead to exploration problems of unsafe behavior \cite{1-3,1-4}. Furthermore, \cite{1-5,1-6} can avoid unsafe actions, but still cannot well balance task performance and safety. Safe reinforcement learning \cite{1-7,1-8,1-9,1-10} solves this problem, but   the constraints  require domain-specific knowledge. In addition to safety requirements, in   dynamic and complex environments, the dynamic scheduling of unmanned systems requires accurate perceptions of environmental information. Traditional scheduling methods often involve extensive rescheduling and recalculation, which often cannot meet the time constraints of strong real-time problems, and dynamic events may  pose long-term potential risks  \cite{1-11,1-12}.

This paper proposes a collaborative optimization algorithm for emergent incident response based on Safe-MADDPG. An emergent incident response constraint-extraction algorithm   learns constraints from the environment, whose relationship information  is encoded   to accurately perceive environmental information, provide information for autonomous decision-making, and obtain environmental constraint conditions. The RND method is used to encourage agent exploration and guide training. The AVFT method is proposed to optimize the delayed reward problem caused by long-term potential risks generated by dynamic future events. Four  indicators are used to evaluate the proposed method.

Our contributions are as follows.

(1) We propose an emergent incident response constraint-extraction algorithm to learn constraints from the environment, whose relationship information   is encoded   to obtain the constraint matrix $M_{c}$ and   constraint threshold $h_{c}$.

(2) We encourage agent exploration by modifying the reward function to encourage agents to detect new environmental states.

(3) We optimize the problem of reward signal delay or   sporadicity in scenario-based reinforcement learning, which dilutes a signal   over time and   weakly affects states far from the time step when the reward is collected.

(4) We design evaluation indicators for emergent incident response in unmanned warehouses.

\section{RELATED WORK}

\subsection{Emergent Incident Response in Unmanned Systems.}
Emergent incidents often dynamically change, resulting in complex states and difficulty in exploration and learning effective strategies. Anomaly-detection methods are currently  used  to detect small-scale emergencies in   early stages. Shames et al. \cite{2-1} constructed an unknown input observer library for two types of distributed control systems with practical significance, and used it to detect and isolate faults in the network, thus providing infeasibility results for available measurements and faults, and a method to remove faulty agents. Deng et al. \cite{2-2} combined structural learning methods with graph neural networks, and used attention weights to  explain detected anomalies, solving the problem of  detecting abnormal events in high-dimensional time-series data. Goodge et al. \cite{2-3} introduced learning to the local outlier factor method in the form of neural networks to achieve greater flexibility and expressiveness, proposing a graph neural network-based anomaly-detection method called LUNAR that uses information from the nearest neighbors of each node to detect anomalies. To alleviate the problem of missing anomalies caused by their reconstruction  by autoencoder-based anomaly detectors, Dong et al. \cite{2-4} proposed the MemAE network, which uses a memory network to obtain sparse codes from the encoder, and   uses them as queries to retrieve the most relevant memory items for reconstruction, enhancing anomaly-detection performance. 

Dynamic scheduling is the main method for emergency dispatch of sudden events, whose solution   usually involves either traditional   methods such as optimization, heuristic methods, and simulation; or intelligent  methods,  including expert systems, neural networks, intelligent search algorithms, and multi-agent methods. Mukhopadhyay et al. \cite{2-7} proposed a two-layer optimization framework to solve a personnel patrol allocation problem in a police department, including a   linear programming patrol response formula to solve the complex and constantly changing distribution of criminals in time and space. Silva et al. \cite{2-8} proposed an emergency service coverage model for sudden events,   considering different priorities   and using priority queueing theory. Mukhopadhyay et al. \cite{2-9} proposed an online event prediction mechanism for the dispatch of emergency personnel for traffic accidents, fires, distress calls, and crime services, and   an algorithm   for the reasonable allocation of   personnel.  
\subsection{Reinforcement Learning and Safety Reinforcement Learning}
The safety of agents is more important than rewards for some reinforcement learning  applications. Most   work related to constraints uses instructions to induce planning constraints \cite{2-12}. Chow et al. \cite{2-18} proposed a primal-dual gradient method for risk-constrained reinforcement learning,   taking policy gradient steps on a trade-off objective between reward and risk while learning a balancing coefficient (dual variable). Constrained policy optimization (CPO) updates the agent’s policy under trust region constraints to maximize its reward while complying with   safety constraints \cite{2-15}. Primal-dual methods solve the difficulty of constrained optimization in each iteration of CPO    \cite{2-16}. These methods  perform well in terms of safety, but poorly in terms of rewards \cite{2-16}. A  class of algorithms for solving constrained Markov decision processes  uses theoretical properties of Lyapunov functions \cite{2-17} in safety value iteration and policy gradient procedures. In   multi-agent safety reinforcement learning, CMIX \cite{2-19} extends QMIX \cite{2-20} by modifying the reward function to consider peak constraint violations and multi-objective constraints.

In the real world, reward signals are often not dense enough, leading to difficulties in reinforcement learning training. In many practical tasks, such as the game of Go and automated chemical design   \cite{2-21}, the final reward or return value can only be obtained after the task is completed, in what is sometimes known as episodic reinforcement learning. When reward signals are delayed or even sporadic, most   deep reinforcement learning algorithms   suffer from poor performance and low sample complexity \cite{2-22,2-23} during  training, in what is known as the credit assignment problem  \cite{2-24}. Rewards can also be delayed or periodic, with nonzero reward values   only obtained when the task ends. For this type of  problem, the reinforcement signal does not immediately appear after the action that triggers the reward, and the algorithm faces the problem of delayed rewards, which is  manifested in the form of poor training performance because of the ambiguous definition of which action triggers the reward  \cite{2-25}. Static reward shaping uses a reward function whose return does not change with the agent’s experience. Before training, a standard reward is provided based on domain knowledge, which is static during training \cite{2-26}. When tasks or environments change, the reward function  should be modified to make it less applicable to multiple environments \cite{2-27}. Dynamic reward shaping is closely related to the proposed method, as it uses a reward function whose time-varying return depends on experience. As the learning process progresses, the reward function is generated based on the experience observed during  training. Marom et al. \cite{2-31} proposed a Bayesian reward-shaping method to allocate rewards, using prior beliefs about the environment as an introductory form to shape rewards. Marthi \cite{2-32} proposed two algorithms to recombine reward functions, aiming to reduce the gap between rewards and actions. Hybrid reward decomposition in multi-objective tasks was evaluated   \cite{2-33}. Effective sampling is important to improve exploration work, and there are already some methods to improve sampling efficiency.

\section{PRELIMINARIES AND PROBLEM STATEMENT}
We introduce the formal definition and key concepts of emergent incident response for unmanned warehouse inbound and outbound operations, and   define the problem of emergent incident response for a multi-agent unmanned warehouse.

\subsection{Concept Definition}

\textbf{Definition 1} (Physical Space Graph of Unmanned Warehouse). The physical space of the unmanned warehouse in this paper is abstracted into a graph $\mathcal{G}=(V, E)$ according to important areas in the real scene. $V$ is a set of nodes abstracted from important hubs or areas that must perform inbound and outbound tasks, such as shelves, inbound areas, and outbound areas. The attribute $w_{i}$ of $V$ represents the number of goods stored at node $v_{i}$.   $V$ is divided into two sets: $V^{n}$,   where no emergent incident occurs, and $V^{f}$,   where one does occur, i.e., $V=\{V^{n},V^{f}\}$ . $E$ is the set of edges between nodes, i.e., the connectivity relationship between key areas such as hubs and shelves where inbound and outbound operations are performed. An attribute $d_{(i,j)}$ of $E$  represents the distance between nodes $v_{i}$ and $v_{j}$.

\begin{figure*}[htp]
    \centering
    \includegraphics[width=18cm]{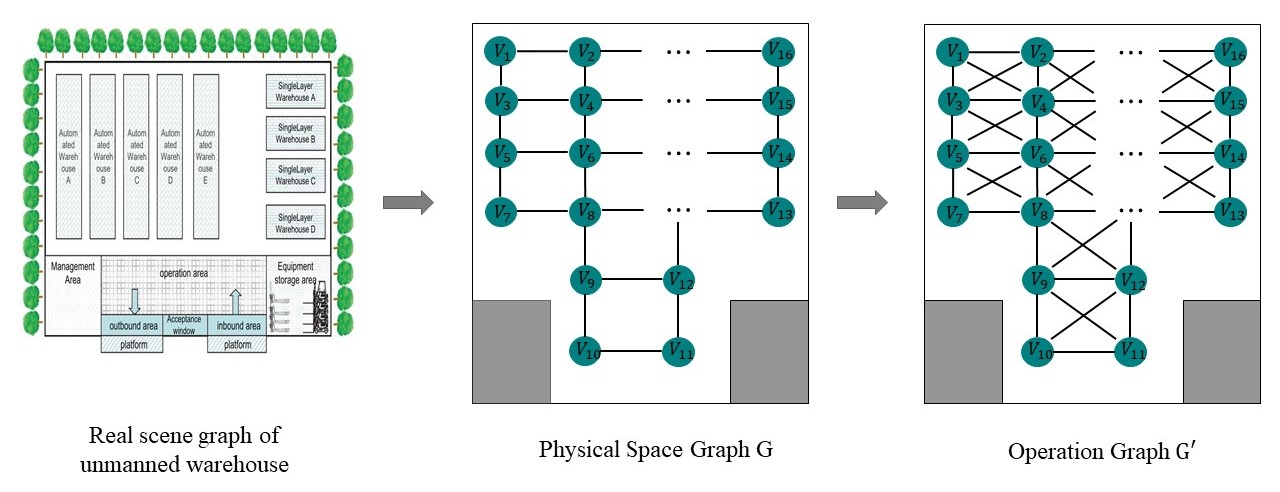}
    \caption{Abstract graph of unmanned warehouse}
    \label{fig:3.1}
\end{figure*}

\textbf{Definition 2} (Unmanned Warehouse Operation Graph). 
  Based on the important areas in the physical space of the unmanned warehouse in the real scene, its physical space graph   $\mathcal{G}$ is obtained. The operation graph   $\mathcal{G}^{'}=(V^{'},E^{'})$ is generated  according to the motion area of the device when performing inbound and outbound operations, as shown in Fig. \ref{fig:3.1}.   $V^{'}$ is the set of device motion area nodes generated based on the node set $V$, and $\mid V \mid $ has no direct relationship with $\mid V^{'} \mid$. The vertex $v^{'}$ has two attributes:   $id$ represents the motion area of a device  in the current device motion area node, and $f(t)$ is the loss caused by emergent incidents to the surrounding environment. Each edge $e_{(i,j)}^{'}$ in $E^{'}$ connects two device motion area nodes, and $\mid E \mid$ has no direct relationship with $\mid E^{'} \mid$.

\textbf{Definition 3} (Sequence of Inbound and Outbound Tasks). The inbound and outbound operation sequence   of the   warehouse is a set of   operations, $\mathcal{O}=\{o_{1}(v_{1}^{'},t_{1}^{d},t_{1}^{e},t_{1}^{n},ga_{1},num_{1}),...,o_{m}(v_{m}^{'},t_{m}^{d},t_{m}^{e},t_{m}^{n},ga_{m},$   $num_{m})\}$,  where $v_{i}^{'}$ is the motion area required to complete   operations; $t_{i}^{d}$ is the deadline of     $o_{i}$; $t_{i}^{e}$ is the expected completion time of  $o_{i}$; $t_{i}^{n}$ is the time already spent on   $o_{i}$; $num_{i}$ is the number of goods  to be transported for $o_{i}$; and $ga_{i}$ is the destination address of  $o_{i}$. This set of   operations can be represented by the value of $t_{i}^{e}$ as a single queue. When performing   $o_{i}$, the device moves quantity $num_{i}$  of goods in the area of node $v_{i}^{'}$  to   destination address $ga_{i}$.

\textbf{Definition 4} (Emergent Incident). An emergent incident is an extraordinary, sudden, and immediately dispositional event that may affect the efficiency of   operations,  which must be immediately resolved.  We denote this  as   $Em=(St,Loc,Type,P_{Type (K)})$, with start time $St$, which occurs at node $Loc$, where $Type$ is the event type, $a$ represents device failure, $b$ represents shelf anomaly, $c$ represents emergent incidents that cause damage to the surrounding environment, and $P_{Type (K)}$ is the probability of severity  $K$ \cite{3-1}. In this paper, $K$ has three levels: $\Rmnum{1}$ (no impact), $\Rmnum{2}$ (may cause impact), and $\Rmnum{3}$ (causes impact), and is calculated as

\begin{equation}
P_{i}(k)=\dfrac{exp(\alpha_{k}+\beta_{k}X_{ki})}{\sum_{\forall k}exp(\alpha_{k}+\beta_{k}X_{ki})},\\
\label{eq:3.1}
\end{equation}
where $\alpha_{k}$ is a constant severity parameter, $\beta_{k}$ is an estimable parameter vector of   severity $K$, and $X_{ki}$ is an explanatory variable vector of   severity category $K$ of   $Type$, such as the running speed of a device,   weight of   transported goods, and environmental conditions.
If  $P_{Type(\Rmnum{1})} \leq P_{Type(\Rmnum{2})}$ or $ P_{Type(\Rmnum{1})} \leq P_{Type(\Rmnum{3})}$ , then  $Em$ occurs at node $Loc$, and   is added to the set of nodes where emergent incidents occur,  $V^{f}\xleftarrow[]{+}\{Loc\}$.

\textbf{Definition 5} (Type of Emergent Incident). Emergent incidents have different constraints. Type $a$, device failure,  includes collision protection failure, communication interruption protection failure, and assembly distance exceeding protection failure. This  manifests as automatic parking during operation. When dealing with such   incidents, it is necessary to release the device from the abnormal state, restore its operation, and ensure that it executes  tasks $o_{i}$, completes   tasks before   deadline $t_{i}^{d}$, and avoids other running devices,  to avoid affecting the normal operation of the   warehouse. Type $a$   incidents have time, space, and resource constraints. Type $b$, shelf anomaly,  includes abnormal cargo posture and shelf collapse, and is  manifested as cargo falling, which affects the device’s execution of   operations and reduces the efficiency of  warehouse operation. When dealing with such emergent incidents, it is necessary to clear the fallen cargo and restore the smoothness of congested roads. Type $b$   has space   and resource constraints. Type $c$, which causes damage to the surrounding environment, includes fires caused by logistics robots, and other devices.  The agent should deal with emergent incidents in a timely manner to avoid further damage to   surroundings. Type $c$   has time, space, and resource constraints.

\textbf{Definition 6} (Constraints on Emergent Incident). Constraints for emergency response to sudden events  consist of time, space, and resource constraints.

Time: After the agent successfully deals with an emergent incident, the device completes     operations before their deadlines.   The time constraint for emergent incidents is

\begin{equation}
t_{i}^{e} + t_{i}^{m} \leq t_{i}^{d} - t_{i}^{n},
\label{eq:3.2}
\end{equation}
 where $v_{i}^{'}$ is the motion area required to complete the   operation, $t_{i}^{d}$ is the deadline of $o_{i}$, $t_{i}^{e}$ is the expected completion time of   $o_{i}$, $t_{i}^{n}$ is the time already spent on   $o_{i}$, and $t_{i}^{m}$ is the time spent on the   response task at   completion node $v_{i}^{'}$.

Space: The safe distance between the agent and other devices  for emergent incident response is

\begin{equation}
\sqrt{(x_{loc_{t}^{i}} - x_{loc_{t}^{j}})^{2} + (y_{loc_{t}^{i}}-y_{loc_{t}^{j}})^{2}} \geq d,
\label{eq:3.3}
\end{equation}
where $x_{loc_{t}^{i}}$ and $y_{loc_{t}^{i}}$ are the positions of device i on the  $X-$ and $Y-axis$, respectively, at time $t$; $x_{loc_{t}^{j}}$ and $y_{loc_{t}^{j}}$ are the respective positions of device $j$ on the $X-$ and $Y-axis$ at time $t$; and $d$ is the minimum safe distance between devices.

Resource: Resources and devices required for the agent to perform emergent incident response tasks   are limited. Hence it is necessary to ensure that these are available.

\textbf{Definition 7} (Extraction of Safety Constraints for Emergent Incident). 
The algorithm to extract constraints for emergent incident response should be pretrained. The agent can obtain constraint information in a timely manner when an   incident occurs, ensuring that the   warehouse can be operated normally, with a timely response to  the   incident. The constraint matrix $M_{c}$ and constraint threshold $h_{c}$ can be obtained through the algorithm. $M_{c}$     is a two-dimensional constraint matrix that represents the constraint entities related to the agent. It  has a binary value, where 1 indicates the presence of a constraint, and 0 indicates its absence. $h_{c}$ is the number of times   the constraint allows an unsafe state to exist.

\textbf{Definition 8} (Emergent Incident Response). 
An emergent incident response   refers to the agent performing   response types $a$ and $b$ after an incident occurs, improving the completion of   emergent incident response and     inbound and outbound operations. For   incident type $c$, the agent performs a response, reduces the damage to the surrounding environment, and prevents the expansion of an incident. In incident $Em$, the probability of no impact, i.e., the maximum probability $P_{Type}(\Rmnum{1})$, is considered a successful  response. For example, when  emergent incident  $em=(st,v_{i},type,P_{type}(K))$ occurs in   operation graph  $G_{'}$, when $P_{Type}(\Rmnum{1}) > P_{Type}(\Rmnum{2})$ and $P_{Type}(\Rmnum{1})>P_{Type}(\Rmnum{3})$ are satisfied, then $V^{f}\xleftarrow[]{-}\{v_{i}\}$.

\textbf{Definition 9} (Emergent Incident Response Agent). The agent is   responsible for     response tasks. Let $A=\{a_{1},⋯,a_{n}\}$ be the set of agents, where an agent moves on the operation graph $G_{'}$ and performs   response tasks that occur at its nodes. After a sudden event occurs, the agent starts an emergency response task from the initial position, ensuring that the equipment completes the  operation tasks before   deadline  $t_{i}^{d}$. The agent must maximize the efficiency of   operations and minimize   losses caused by  incidents without interrupting  operations. The problem proposed in this paper is real-time. At time $t$, if an  incident occurs at a node, the set of emergent incident nodes $V^{f}$ is updated. The agents complete the   response tasks through communication and collaboration. The execution time of an agent’s task is inversely proportional to the number of agents.   The time required to complete response task nodes and the time spent by $n$ agents to complete the   task at node $v_{i}^{'}$ is

\begin{equation}
t_{i}^{m}=\dfrac{CT}{n}.
\label{eq:3.4}
\end{equation}

\subsection{Multi-agent Emergency Decision-making}
A model is established to optimize emergent incident response in   inbound and outbound operations. An undirected graph $\mathcal{G}=(V, E)$ is generated based on the real unmanned warehouse scenario to represent its physical space graph, based on which   an operation graph of the unmanned warehouse, represented by an undirected graph $\mathcal{G}^{'}=(V^{'},E^{'})$, is generated, as shown in Fig. \ref{fig:3.1}. $V^{'}$ is a collection of device motion area nodes generated based on   node set $V$ of     $\mathcal{G}$. The vertex $v^{'}$ has   attribute $id$, indicating that the current device motion area node is the motion area of device $id$. Each node   contains the attribute $w_{i}$,   representing the number of goods placed at node $v_{i}^{'}$ when $w_{i}>0$, and   $w_{i}=0$ indicates that no goods are placed at node $v_{i}^{'}$. The attribute $f(t)$ represents the loss caused by the emergency incident to the surrounding environment. $E_{'}$ contains each edge $e_{i,j}^{'}$ that connects     nodes $v_{i}^{'}$ and $v_{j}^{'}$. The attribute   $E_{'}$ in $d_{i,j}^{'}$  represents the distance between nodes $v_{i}^{'}$ and $v_{j}^{'}$. $A=\{a_{1},⋯,a_{n}\}$ is a given set of $n$ agents, and the initial inbound and outbound operation tasks    $\mathcal{O}=\{o_{1}(v_{1}^{'},t_{1}^{d},t_{1}^{e},t_{1}^{n},ga_{1},num_{1}),...,o_{m}(v_{m}^{'},t_{m}^{d},t_{m}^{e},t_{m}^{n},ga_{m},$   $num_{m})\}$ are given, where $v_{i}^{'}$ is the motion area required to complete the  operation, $t_{i}^{d}$ is the deadline of   $o_{i}$, $t_{i}^{e}$ is the expected completion time of   $o_{i}$, $t_{i}^{n}$ is the time already spent on   $o_{i}$, $num_{i}$ is the number of goods that must be transported for   $o_{i}$, and $ga_{i}$ is the destination address of $o_{i}$. When there is no emergent incident, the devices in the   warehouse transport goods from   area node $v_{i}^{'}$  to the destination address  before the deadline. When an   incident occurs,   agents and devices handle it without conflict, maximizing the efficiency of the   operation  during the  response process, i.e., $max\sum\dfrac{1}{T_{ime}(O)}$, and the loss caused by the   incident is minimized, i.e., $min\sum V_{alue}(V^{f})$, $T_{ime}(O)$, where $T_{ime}(O)$ is the total time for the devices to complete the   tasks, and $V_{alue}(V^{f})$ is the total  loss caused by the   incident.

\section{PROPOSED SOLUTIONS}
\subsection{Constraint-extraction algorithm}
In the   dynamic and complex emergent incident response environment of an unmanned warehouse, it is necessary to obtain constraint information of   agents during the   response process as well as the dynamic environment to reduce   safety risks  and avoid chain accidents. Based on the obtained constraint information, the agents are trained to avoid making decisions that have a negative impact on the target during the training process. Constraints are traditionally obtained  based on   conditions designed by domain experts; domain professionals are usually required  to design these, and they cannot consider all constraints in the environment. We designed a constraint-extraction algorithm   based on massive simulation data, which can more comprehensively represent   existing constraints in the current environment,      and enable agents to make decisions that meet constraints and achieve goals.

\begin{figure*}[htp]
    \centering
    \includegraphics[width=16cm]{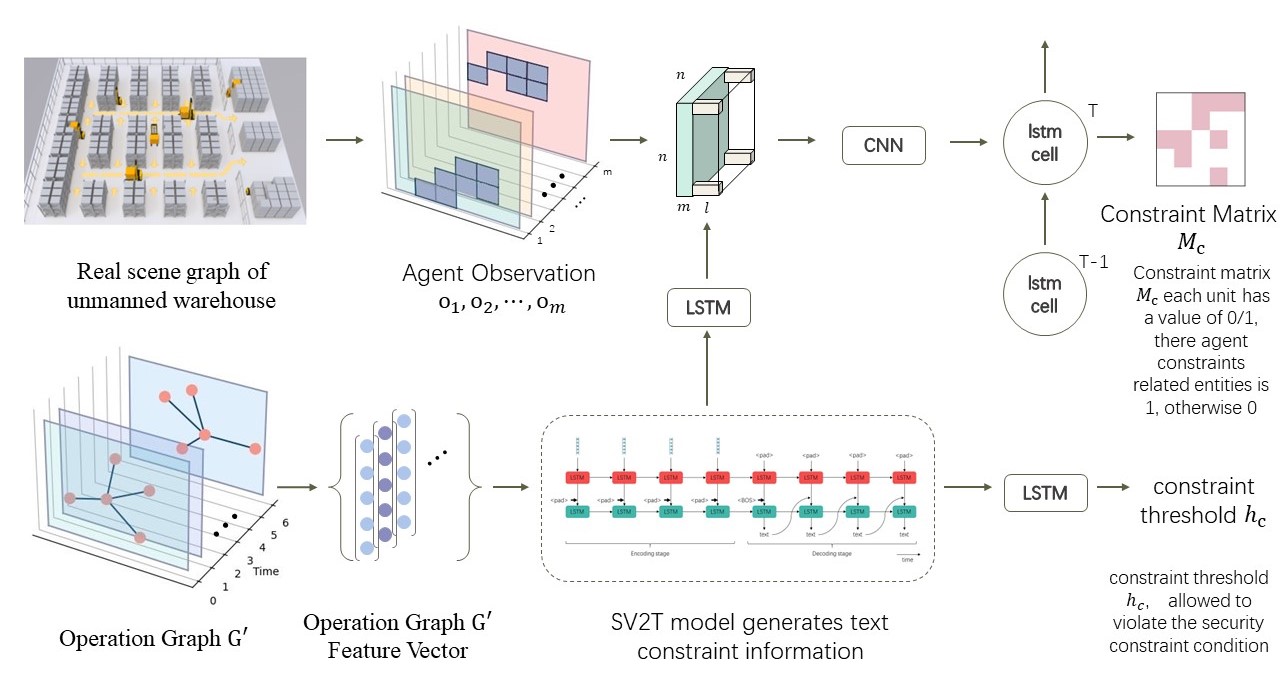}
    \caption{Structure graph of constraint-extraction algorithm for emergent incident response}
    \label{fig:5.1}
\end{figure*}

The spatial information between nodes in     operation graph $\mathcal{G}^{'}$ is obtained through the graph neural network GCN. The constraint information changes continuously with time, and traditional constraints are mostly static, which cannot accurately represent the constraint information in the environment. We propose a constraint-extraction algorithm, whose structure     is shown in Fig. \ref{fig:5.1}, that uses the S2VT model and LSTM to obtain sequence information that changes dynamically over time.  We encode the collection of   operation graphs $\{\mathcal{G}_{1}^{'}, ..., \mathcal{G}_{t}^{'}\}$ through the   GCN, and generate the text information $text$ of the constraints through the S2VT model \cite{3-2}, such as the safety distance between agents. The embedding of size $l$ obtained through the LSTM neural network is connected with the observation feature vector of size $n×n×m$ of $m$ agents, and the embedding vector is obtained through the CNN. The constraint matrix $M_{C}$ is obtained through this embedding vector, and has size $n×n$, where each unit value is 0 or 1. A value of 1 indicates that there is an entity related to the agent constraint in the corresponding unit of the observation value, and a value of 0 indicates that there is no such entity. The order constraint   of the safety status of the agent and   past events has a long-term dependence, and $M_{C}$ depends on the past status accessed by the agent. An LSTM layer is added before calculating $M_{C}$ to consider the past status. In addition, we input the text information $text$ to the LSTM neural network to obtain a scalar $h_{C}$, which represents the actual scalar of the constraint threshold, i.e., the number of times a safety constraint state is allowed to be violated. Table \ref{algorithm2} shows the main steps of the constraint-extraction algorithm, whose training     includes that of   constraint matrix $M_{C}$ and   constraint threshold $h_{C}$. For  $M_{C}$, we use the cross-entropy loss function for training,

\begin{equation}
Loss(M_{C})=-\sum_{(x_{t},G^{'})\sim D_{data}}[\dfrac{1}{|M_{c}|}\sum_{i,j=1}^{n}zlog\hat{z}+(1-z)log(1-\hat{z})],
\label{eq:3.9}
\end{equation}
where $z$ is the target constraint matrix $\hat{M}_{C}$, defined as the binary mask label of the $i$-th row and $j$-th column of the observation value set $X$ of $n×n$ agents, and $\hat{z}$ is the constraint matrix $M_{C}$ output by the emergent incident response algorithm.

We use the mean squared error (MSE) loss function, 

\begin{equation}
Loss(h_{C})=\sum_{(x_{t},G^{'})\sim D_{data}}[(h_{c}-\hat{h}_{c})^2],
\label{eq:3.10}
\end{equation}
 to train the constraint threshold $h_{c}$.
 
The dataset for training the constraint-extraction algorithm   comes from real-world emergent incident log files   provided by logistics companies, and   log files obtained from simulations. The emergent incident log files record the entire process data of   emergent incidents. The data are converted to the complete process of an emergent incident, represented by a set of sequential unmanned warehouse operation graphs $G^{'}$, from which the   operation graph $G^{'}$ dataset for training the constraint-extraction algorithm   is generated.

\begin{algorithm}
    \caption{Constraint-extraction   for emergent incident response}
    \label{algorithm2}
    \begin{algorithmic}[1]  
        \REQUIRE Unmanned warehouse operation graph $G^{'}$,   emergent incident response agent observations $X=\{x_{1},...,x_{n}\}$.
        \ENSURE Constraint matrix $M_{C}$, Constraint thresholds $h_{c}$.
        \WHILE{$t<$ end time $T$}
            \STATE Obtain   observations of   emergency response agent $X$.
            \STATE The $n×n$ observed values of $m$ emergency response agents are connected as  $n×n×m$ matrix $F_{X}$.
            \STATE Get unmanned warehouse operation graph $G^{'}$ from known physical space graph $G$ of unmanned warehouse.
            \STATE Get $G^{'}$ feature vector: $GCN(G^{'} \rightarrow F_{G^{'}})$.
            \STATE $F_{G^{'}}$ is passed through   $S2VT$ model to generate   constraint text information $C_{text}$
            \STATE $LSTM_{\theta}(C_{text}) \rightarrow F_{C_{text}}$.
            \STATE Constraint matrix $M_{C}$: $LSTM$     $Cell(CNN(F_{X},$      $F_{C_{text}})) \rightarrow M_{C}$
            \STATE Constraint thresholds $h_{c}$: $LSTM_{\theta^{'}}(C_{text}) \rightarrow int(h_{c})$
            \STATE Equation \ref{eq:3.9}, Calculating constraint matrix $M_{C}$ loss function and update parameter.
            \STATE Equation \ref{eq:3.10}, Calculating constraint thresholds $h_{c}$ loss function and update parameter.
            \STATE $t \leftarrow t+1$
        \ENDWHILE
    \end{algorithmic}
\end{algorithm}
\subsection{CMDP Problem Construction}
We transform the optimization problem of emergent incident response to a constrained Markov decision process with $n$ agents, $A=\{a_{1},⋯,a_{n}\}$, consisting of a tuple $(S,U,\pi,R,\gamma)$. The Markov decision process of $n$ agents is defined as a state set $S$,  local observation set $X=\{x_{1},⋯,x_{n}\}$, and   action set $U=\{u_{1},⋯,u_{n}\}$.   $S$ includes a constraint matrix $M_{c}$,  constraint threshold $h_{c}$,  set of inbound and outbound operations $O$, a set of completion statuses of inbound and outbound operations $\omega(t)$,  set of emergent incident nodes $V_{t}^{f}$,   current node position $loc_{i}(t)$ of the agent, and   target node position $g_{i}(t)$ of the agent. X is the local observation space of the agent, whose observation value   at time $t$ is a part of   state $S$, $x(t) \in S$. $U$ is the action set of the agent. For a given state  $s \in S$, the agent selects an action from the action space based on the observation results through a policy, i.e., $\pi:S \leftarrow U$.

\subsubsection{Environment State}
The state $S(t)$ at time $t$ is represented by a tuple $\{M_{c}, h_{c}, I_{t}, A_{t}\}$, where $M_{c}$ is the constraint matrix, $h_{c}$ is the constraint threshold, $I_{t}$ is the relevant information of inbound and outbound operations, and $A_{t}$ is the information of the agent at time $t$. The attributes of $I_{t}$ include $\{O,\omega(t),V_{t}^{f}\}$, where $O$ represents the set of inbound and outbound operations, $\omega(t)$ is the completion status of   $O$ at time $t$, and $V_{t}^{f}$ is the set of emergent incident nodes at time $t$. $\mid A(t) \mid=\mid A \mid$, whose each element $a_{i}(t) $ corresponds to a set of attributes $\{loc_{i}(t),g_{i}(t)\}$, where $loc_{i}(t)$ is the current node position of agent $a_{i}$, whose  target node position is $g_{i}(t)$.

\subsubsection{Action}  
In the optimization problem of emergent incident response, the actions of the agent are divided into three types: (1) for emergent incident type $A$,   equipment failure, the agent performs emergent incident response to restore the state of the equipment where the   incident occurred; (2) for   incident type $B$,   shelf collapse, the agent performs the transportation of goods at the node where the emergent incident occurred; (3) for   incident type $C$, equipment fire, the agent performs emergent incident response to the equipment where the   incident occurred. The action at time $t$ is represented by $U(t)$, which consists of a tuple $\{U_{s}(t),U_{c}(t),U_{MOVE}(t)\}$, where $U_{s}(t)$ is the emergent incident response action, $U_{c}(t)$ is the transportation action of the agent for goods at the node where the   incident occurred, and $U_{MOVE}(t)$ is the movement action of the agent. Based on the current observation, the agent selects values from the movement action space $U_{MOVE}(t)$, and selects different emergent incident response actions from $U_{s}(t)$ and $U_{c}(t)$, based on the different types of emergent incidents. The action of the agent at time $t$ and $u(t) \in U$ is

\begin{equation}
U(t)=\{U_{s}(t),U_{c}(t),U_{MOVE}(t)\}.
\label{eq:3.11}
\end{equation}

\subsubsection{Reward}
The reward function has   three parts.  $R_{succ}$ is the reward for successfully handling   incidents, $R_{work}$ is the completion status of inbound and outbound operations, and $R_{loss}$ is the loss caused by   incidents. The importance of each objective is indicated by a weight that increases with importance. $\alpha$, $\beta$, and $\gamma$ are the respective weight coefficients for successfully handling   incidents,   efficiency of completing   operations, and   losses caused by   incidents,  with values in $[0,1]$. If the weight coefficient is 0, the objective is not considered, and if it is 1, the objective is   most important. The objective function is

\begin{equation}
r_{(i,t)}=max\sum_{t=1}^{T}\sum_{i=1}^{N}(\alpha R_{succ}+\beta R_{work}+\gamma R_{loss}). 
\label{eq:3.12}
\end{equation}

 The first objective is to successfully handle emergent incidents, whose reward function   $r_{s}$ has a fixed, discrete value. The reward function for handling failures is  $-r_{s}$. Emergent incidents have different criteria for judgment. For equipment failure   incidents, there are  collision protection failures, communication interruption protection failures, and failures of the assembly distance exceeding values that cause the equipment to automatically stop. The standard for   handling this type of   incident is to resolve the abnormal state of the equipment and allow it to continue running. The standard for successfully handling shelf-collapse   incidents is to remove the fallen goods, which affects the equipment’s ability to perform   operations and reduces the efficiency of the  warehouse.

The second objective is to represent the completion status of  operations, i.e., to successfully transport goods to their destination address  $ga_{i}$ before   deadline $t_{i}^{d}$. The optimal result is to complete as many   operations as possible. The completion status of   operations   is represented by $t_{i}^{d}$,$t_{i}^{e}$,$t_{i}^{n}$,$t_{i}^{m}$. The reward function $R_{work}$ for the completion status of   operations   has a discrete value, with a set fixed value $r_{w}$   for successfully completing  operations, and a reward function   $-r_{w}$ for not completing   them.

  The third objective represents  losses caused by  incidents. If an   incident     does not cause damage, then $R_{loss}$ does not need to be calculated, and its weight coefficient is set to $\gamma$=0. If an   incident occurs, such as equipment catching fire, which can easily   damage   the surrounding environment, the weight coefficient is set according to the importance of the objective, with a value of $\gamma \in (0,1]$. We use the function proposed earlier to represent such losses. The objective function is

\begin{equation}
R_{loss} = \bigtriangledown f(t) \cdot e^{-f(t)}.
\label{eq:3.15}
\end{equation}

\subsection{Collaborative optimization algorithm of emergent incident response based on Safe-MADDPG}

\subsubsection{Encouraging agent exploration}  

We encourage agents to explore using the RND \cite{3-3} method, whose idea  is to quantify the novelty of states based on the prediction error of the network trained on the agent’s past experiences, i.e., the neural network has a lower prediction error on data similar to what it was   trained on, and higher error on dissimilar data.   A larger intrinsic reward is given to encourage exploration for more novel states, and a smaller reward   for less novel states. As the rewards in the environment are sparse, we use exploration reward $r_{t}= e_{t}+i_{t}$, where $e_{t}$ is the external reward, i.e., the original reward from environmental feedback, and $i_{t}$ is the intrinsic reward, which is calculated using   a fixed, randomly initialized target network $f$, and a prediction network $\widetilde{f}$ trained using data collected by the agent. The target network $f$ maps inputs to $f:O \rightarrow R^{k}$, while the prediction network $\widetilde{f}:O \rightarrow R^{k}$ maps inputs to predicted outputs.   $\widetilde{f}$ is trained using a mean squared error (MSE) loss function $e = \parallel \widetilde{f}(x;\theta)-f(x) \parallel^{2}$, whose parameters $θ_{\widetilde{f}}$   are updated by minimizing the loss function using gradient descent. The error   $e$ of   $\widetilde{f}$ is used as the intrinsic reward. According to the   MSE calculation formula, the larger the difference between the input state and   previous state, the larger the value of $e$, indicating that the state is more novel. This encourages the agent to explore more novel states.

\subsubsection{Delay reward optimization} 

To obtain the immediate impact of the current action on the environment, we propose the action value function transformer (AVFT) method to predict its action value function $Q(s,a)$;   better evaluate whether the current action   helps in the completion of the emergent incident response strategy; and for the agent to better handle     incidents   during   training.

We define the expected target reward $\widetilde{R}_{T} =\sum_{t^{'}=t}^{T}r_{t}$, which is different from the expected return $R_{T}=\sum_{t=0}^{T}r_{t}$ in reinforcement learning, and is the cumulative reward of a trajectory.  $\widetilde{R}_{T}$ is the future cumulative reward after executing an action. During   training, the expected target reward is initialized, and after making an action decision, $\widetilde{R}_{T}$ is updated as $\widetilde{R}=\widetilde{R}+[\widetilde{R}[-1]-r]$, where $r$ is the reward value from environmental feedback after making the action decision.

\begin{figure}[htp]
    \centering
    \includegraphics[width=9cm]{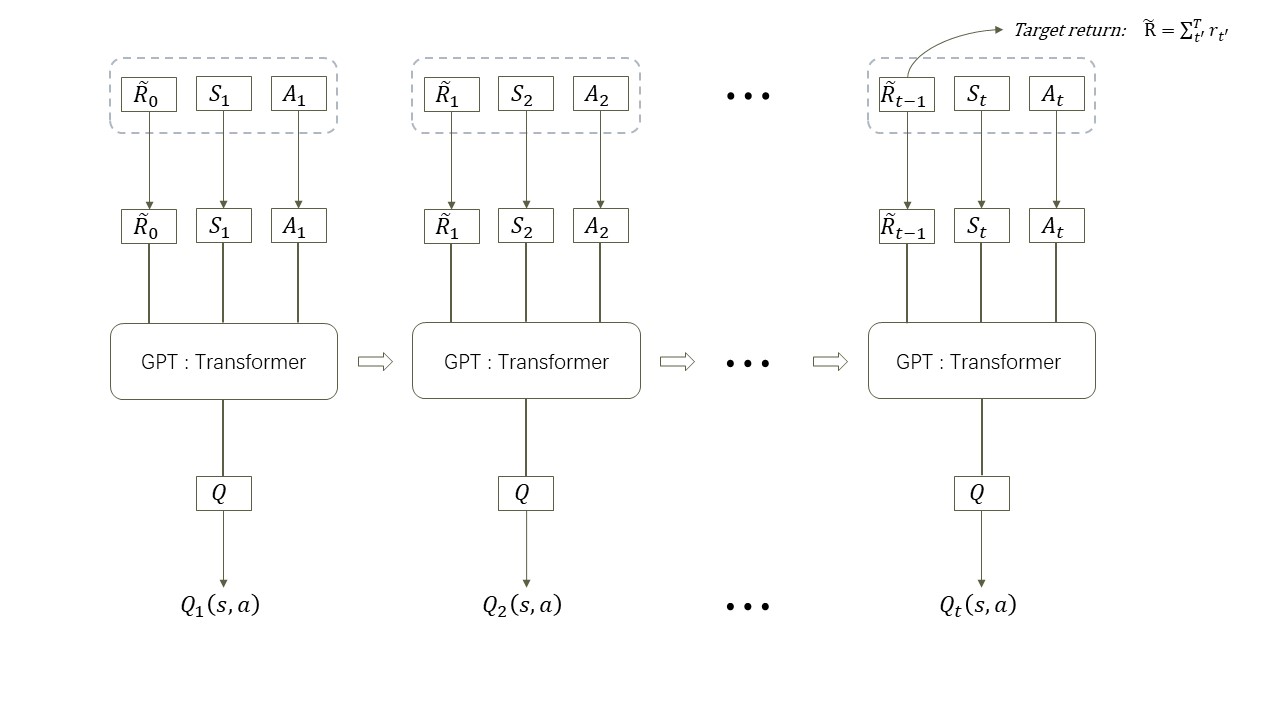}
    \caption{AVFT Method Framework Graph}
    \label{fig:5.3}
\end{figure}

The expected target reward (sum of future rewards), state, and   predicted action value function $Q(s, a)$ are selected as the trajectory representation,  $\tau= (\widetilde{R}_{0},s_{1},u_{1},\widetilde{R}_{1},s_{2},u_{2},⋯,\widetilde{R}_{T-1},s_{T},u_{T})$, where $\widetilde{R}_{T}=\sum_{t^{'}=t}^{T}r_{t}$, $s_{t}$ is the expected target reward, $s_{t}$  is the state of the agent, and $u_{t}$  represents the action of the agent at time $t$. That is, at time $t$,   $Q(s, a)$ is predicted based on the expected reward at time $t-1$, the state $s_{t}$, and the action $u_{t}$ at time $t$. The last $K$ time steps are used as input to the AVFT method, resulting in a total of $3K$ data points (each time step includes the expected reward, state, and action). The original input is projected to the embedding dimension using a linear layer to obtain the token embedding, which is   processed by the AVFT method to calculate $Q(s, a)$, representing the impact of the current action on the expected target reward $\widetilde{R}$. Fig. \ref{fig:5.3} shows the   framework of the AVFT method. The loss function for training it is

\begin{equation}
Loss = (y_{i}-Q_{i}^{u}(S,u;\rho)^{2},
\label{eq:3.16}
\end{equation}
where $y_{i}=R_{i,t}^{'}+\gamma Q_{i}^{'}(S,u^{'})$, $S$ represents the state of the agent, $u$ represents the action of the agent, and $\rho$ represents the parameters of the Q network. For the AVFT method, MSE is used to train the predicted action value function $Q(s, a)$.
\subsubsection{Collaborative optimization algorithm for emergent incident response} 
We introduce a collaborative optimization algorithm for emergent incident response, combining the constraint-extraction algorithm for emergent incident response and the improved Safe-MADDPG multi-agent safety reinforcement learning algorithm to handle   incident responses to nodes in unmanned warehouses.

The constraint-extraction algorithm for emergent incident response is   used to extract the  constraint information of   agents during the emergent incident response process and that of the dynamic environment of the unmanned warehouse on the agents, and incorporate it as part of the information for   agents to make decisions. This reduces   safety risks and prevents chain accidents. Algorithm \ref{algorithm3} summarizes the     process of the collaborative optimization algorithm for emergent incident response based on Safe-MADDPG. Set and parameter initializations are performed in lines 2--4, including initializing the set of unmanned warehouse operations $O$,  sudden event nodes $V^{f}$, and   agent position information $Loc(t)=\{loc_{1}(t), loc_{2}(t),⋯, loc_{n}(t)\}$. In line 7, the initial state $S$ of the environment feedback is obtained, and in line 9, the constraint matrix $M_{c}$ and constraint threshold $h_{c}$ information are obtained from Algorithm \ref{algorithm2}. In line 10, the information of the $\omega(t)$ set of unmanned warehouse operation execution status is obtained from Algorithm, and the agents share information about the emergent incident response related to the surrounding nodes of their current node and set $V^{f}$. In line 11, the matrix $M_{b}$ at step $t^{'}$ is calculated, to calculate the value of violating the constraint at step $t^{'}$ if there is a constraint entity in the corresponding element position $(i, j)$ of   constraint matrix $M_{c}$, and otherwise the element value at position $(i, j)$ of   matrix $M_{b}$ is set to 0.   $M_{b}$ is calculated as

\begin{equation}
M_{b} = (\sum_{t=0}^{t^{'}}\hat{C}(s_{t},a_{t},G^{'})-h_{c}) \ast M_{c}.
\label{eq:3.17}
\end{equation}
In lines 12--15, the agent obtains the safe action set $u_{safe}^{t}$ through the actor network and the safety layer, and executes this   to obtain new observations and rewards. In lines 16--17, the RND method is used to better guide the agent’s exploration with rewards, and   $(S,u_{safe}^{t},R_{t}^{'},S^{'},M_{c},M_{b},h_{c})$   is stored in the experience pool $D$. In line 18, the set $\omega(t)$  of unmanned warehouse operation execution status and $V^{f}$  are updated. In lines 19--30, the actor   and critic networks and their target networks, are updated, as is $\widetilde{f}(S;\theta)$.

\begin{figure*}[htp]
    \centering
    \includegraphics[width=16cm]{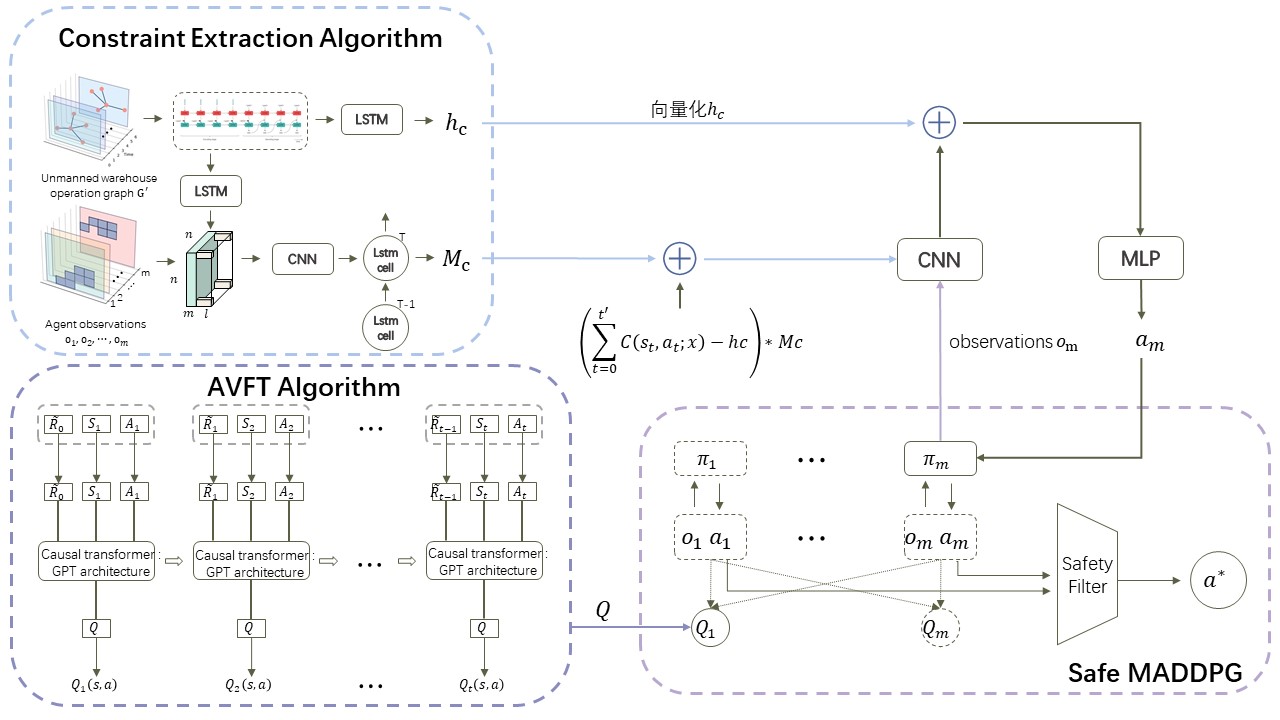}
    \caption{Collaborative optimization algorithm framework for emergent incident response based on Safe-MADDPG}
    \label{fig:5.4}
\end{figure*}

Fig. \ref{fig:5.4} shows the    framework of the collaborative optimization algorithm for emergent incident response based on Safe-MADDPG.
\begin{algorithm}
    \caption{Collaborative optimization   for emergent incident response based on Safe-MADDPG}
    \label{algorithm3}
    \begin{algorithmic}[1]  
        \REQUIRE Unmanned warehouse operation graph $G^{'}$, emergent incident node set $V^f$, inbound and outbound operation execution set $\omega(t)$ .
        \ENSURE Decision-making $u_{t+1}$ of emergent incident response agent at time $t+1$.
        \STATE Pretrained Algorithm \ref{algorithm2}
        \STATE Initialize the set of unmanned warehouse inbound and outbound tasks sequences $O$
        \STATE Initialize emergent incident nodes set $V^f$
        \STATE Initialize the location information of the emergent incident response agent $Loc(t)=\{loc_{1}(t), loc_{2}(t),⋯, loc_{n}(t)\}$
        \FOR{episode k=1 to M} 
            \STATE Initializes random process $N$ to action exploration
            \STATE Feedback initialization state $S$,$Env \rightarrow S$.
            \FOR{t=1 to max-episode-length}
                \STATE $M_{C}$,$h_{c}$ for Algorithm \ref{algorithm2}.
                \STATE $\omega (t)$ for Algorithm \ref{algorithm1}.
                \STATE Calculation $M_{b}$ for equation \ref{eq:3.17}
                \STATE Select the action $u_{i}=u_{θ_{i}}(s_{i},M_{c},M_{b},h_{c})+N_{t}$.
                \STATE $u^{t}=(u_{1},u_{2},⋯,u_{n})$
                \STATE The security layer converts the action $u^{t}$ to the set of security actions $u_{safe}^{t} \leftarrow safety Filter(u^{t})$
                \STATE Execute the safe action set $u_{safe}^{t}$ to get the reward of environment feedback.
                \STATE The RND method calculates the reward function: $R_{t}^{'}=R_{t}+I_{t}$, $I_{t} = \parallel \widetilde{f}(x;\beta)-f(x) \parallel^{2}$
                \STATE Store $(S,u_{safe}^{t},R_{t}^{'},S^{'},M_{c},M_{b},h_{c} )$ to D
                \STATE Update $\omega(t)$, $V^f$
                \STATE D extracts mini-batches $(S,u_{safe}^{t},R_{t}^{'},S^{'},M_{c},M_{b},$  $h_{c})$
                \STATE $u^{'}=(u_{θ_{1}}^{'}(s_{1},M_{c},M_{b},h_{c}),u_{θ_{2}}^{'}(s_{2},M_{c},M_{b},h_{c}),...,$   $u_{θ_{n}}^{'}(s_{n},M_{c},M_{b},h_{c}))$
                \STATE The AVFT method calculates $Q(S,u)$
                \STATE $z_{i}=R_{i,t}^{'}+\gamma Q_{i}^{'}(S,u^{'};\rho)$
                \STATE Update the critic network, $L(θ_{i})=\dfrac{1}{mini-batch}\sum(z_{i}-Q_{i}^{u}(S,u;\rho))^{2}$
                \STATE Update the actor network, $\bigtriangledown_{θ_{i}}J_{i}=\dfrac{1}{mini-batch}\sum \bigtriangledown_{\theta_{i}}u_{i}(s_{i})\bigtriangledown_(u_{i} )Q_{i}^{u}(S,u;\rho)$
                \STATE Update $\widetilde{f}(S;\theta)$, $Loss=\parallel \widetilde{f}(x;\beta)-f(x) \parallel^{2}$
                \STATE Update target network, $\widetilde{\theta_{i}} \leftarrow \tau \theta_{i} + (1-\tau)\widetilde{\theta_{i}}$, $\widetilde{\rho_{i}} \leftarrow \tau \rho_{i} + (1-\tau)\widetilde{\rho_{i}}$.
            \ENDFOR
        \ENDFOR
    \end{algorithmic}
\end{algorithm}

\section{EXPERIMENTAL RESULTS AND DISCUSSION}
We performed simulation experiments with different  parameters   to test the performance of the system and verify its effectiveness. 
\subsection{Evaluation Index}
\subsubsection{Reward} 
The   reward function is
\begin{equation}
R=\dfrac{\sum_{i=1}^{m}r_{i}}{m},
\label{eq:3.18}
\end{equation}
where $r_{i}$ is the total sum of rewards that   agent $i$ receives from the environment after one round of training, and $m$ is the number of agents. This   represents the average   total rewards obtained by all agents in the training round.

\subsubsection{Completion Rate of Unmanned Warehouse Operations} 
The completion rate of unmanned warehouse operations is the ratio of the number of completed warehouse operations to the total number of warehouse operations, reflecting the impact on   efficiency of emergency response to sudden events. An effective model strategy  will have a positive impact on the completion rate of   operations,

\begin{equation}
rate_{s}=\dfrac{\sum_{i=1}^{N}n_{i}}{n_{O} \ast N},
\label{eq:3.19}
\end{equation}
where $n_{i}$ is the number of completed warehouse operations by the $i$-th automated device, $n_{O}$ is the total number of warehouse operations, and $N$ is the number of automated devices performing warehouse operations.

\subsubsection{Loss Rate of Emergent Incidents} 
  If the model’s strategy is effective, the loss rate of emergent incidents,

\begin{equation}
rate_{f}=\dfrac{n_{v_{f}}}{n_{V}},
\label{eq:3.20}
\end{equation}
will be relatively small, where $n_{v_{f}}$ is the number of nodes where emergent incidents occur, and $n_{V}$ is the total number of nodes in the environment.

\subsubsection{Completion Rate of Emergent Incident Response} 
The completion rate of emergent incident response   represents the successful handling of emergent incidents, and is defined as
\begin{equation}
rate_{sf}=\dfrac{n_{v_{f}}^{'}}{n_{v_{f}}},
\label{eq:3.21}
\end{equation}
where $n_{v_{f}}^{'}$ is the number of nodes where emergent incident response has been completed, and $n_{v_{f}}$ is the number of nodes where emergent incidents occur.

If the model’s strategy is effective, this will be relatively high.

\subsection{Experiment Scene of Unmanned Warehouse Operation}
\begin{figure}[htp]
    \centering
    \includegraphics[width=9cm]{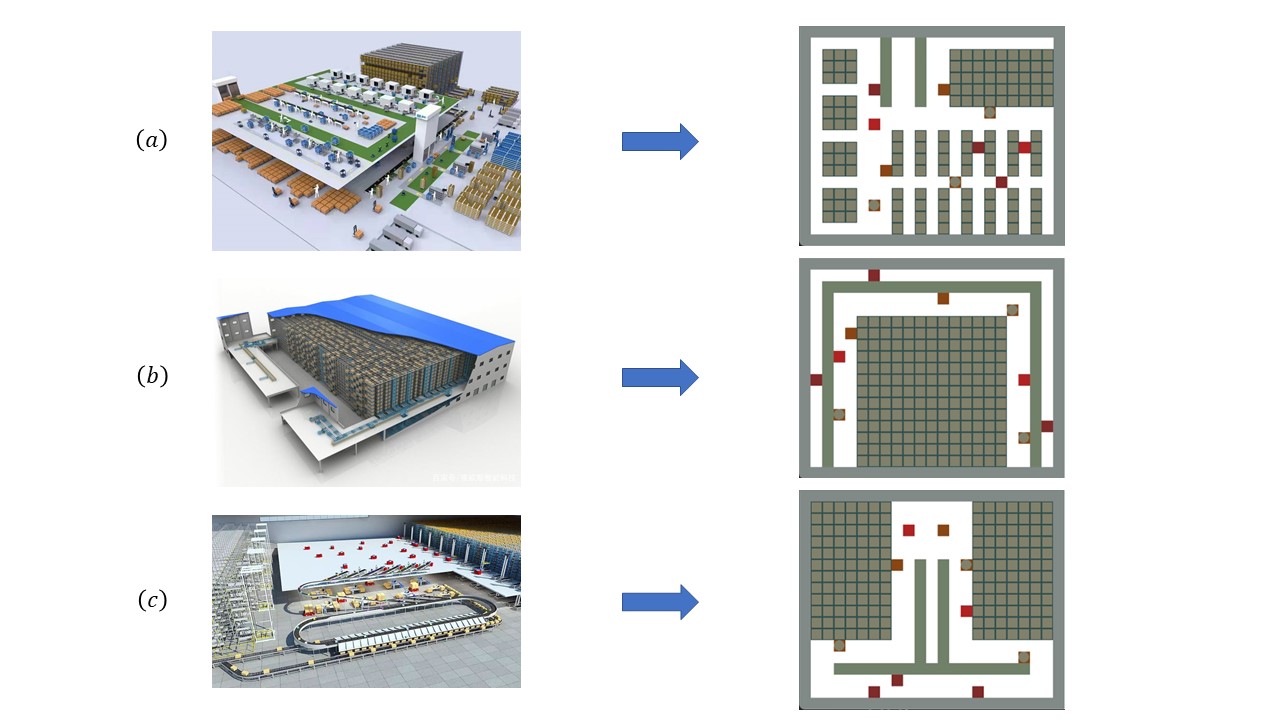}
    \caption{Experimental plan of unmanned warehouse}
    \label{fig:6.1}
\end{figure}

We conducted experiments on unmanned warehouse scenarios with three   layouts. Due to the large storage area of such warehouses, different numbers of emergent incidents were set. The   configurations were: 2 emergent incident response agents and 4 emergent incidents, 3 emergent incident response agents and 6 emergent incidents, and 4 emergent incident response agents and 8 emergent incidents. As shown in Fig. \ref{fig:6.1}, according to the   layouts of unmanned warehouse scenarios, we designed three   layout types of unmanned warehouse simulation scenarios, referred to as layouts $A$,  $B$, and   $C$.

\subsection{Analysis of Experimental Results}
\begin{figure}[htp]
    \centering
    \includegraphics[width=9cm]{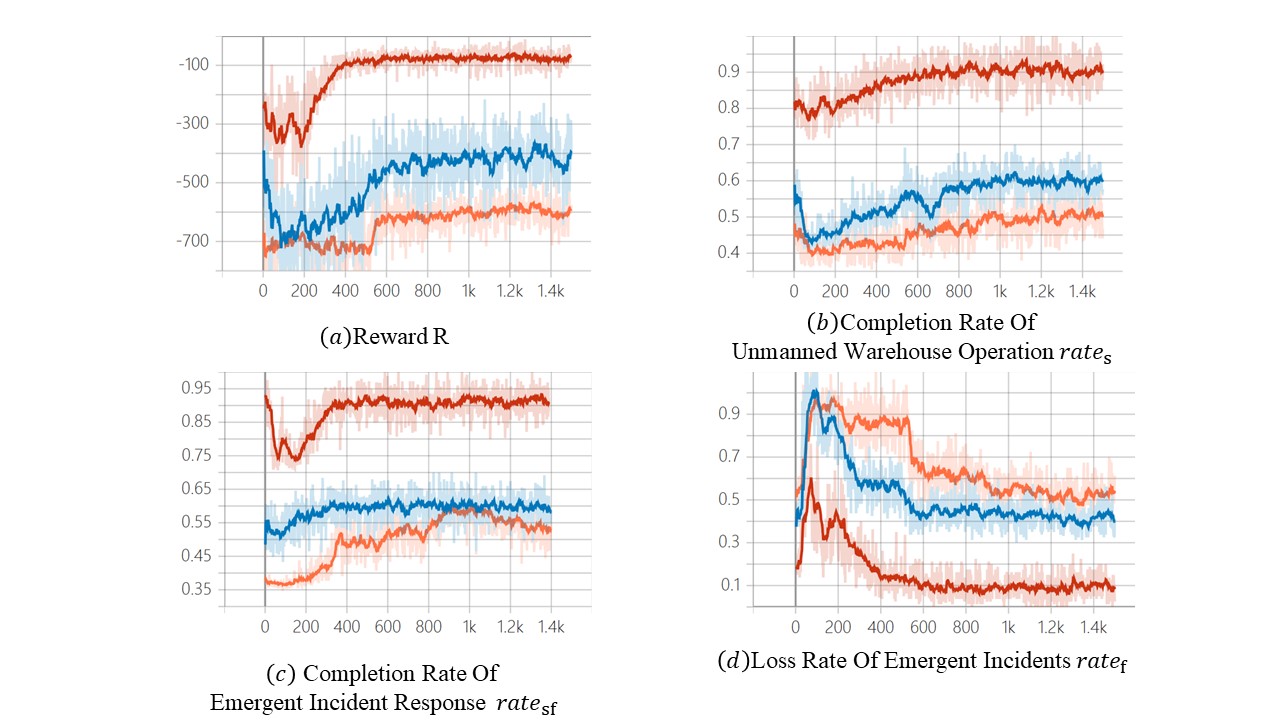}
    \caption{Experimental results}
    \label{fig:6.2}
\end{figure}

We take   layout $A$, with 3 agents for emergent incident response and 6 emergent incidents, as an example. The experiment uses the EI-Safe-MADDPG algorithm based on Safe-MADDPG for emergent incident response coordination optimization, the C-Safe-MADDPG algorithm with emergent incident response constraints, and the Safe-MADDPG algorithm to train the multi-agent system for emergent incident response. After 20,000 training episodes, the experimental evaluation indicators are averaged every 15 training episodes, and the results are shown in Fig. \ref{fig:6.2}, where $(a)$ represents the reward function $r$; $(b)$   represents the completion rate $rate_{s}$ of inbound and outbound operations; $(c)$ represents    the emergent incident response completion rate $rate_{sf}$ of inbound and outbound operations; and $(d)$   represents the emergent incident response loss rate $rate_{f}$ of inbound and outbound operations.

Fig.  \ref{fig:6.2}(a) shows the results of the reward function. It can be seen that, compared to   Safe-MADDPG, the C-Safe-MADDPG algorithm has a significant improvement in the reward function, while   EI-Safe-MADDPG   converges faster and has a greater improvement. In Fig.  \ref{fig:6.2}(b), it can be seen that among all   algorithms,   EI-Safe-MADDPG   has the highest completion rate  $rate_{s}$ for inbound and outbound operations, which can reach 92.7\%. The completion rate of C-Safe-MADDPG   is second-highest, with a maximum of 59.5\%, while  Safe-MADDPG  has the worst result, at around 48.5\%. The completion rate of EI-Safe-MADDPG   converges faster and more steadily compared to other algorithms, and the   completion rate after convergence is   the highest. From Fig.  \ref{fig:6.2}(c), it can be seen that   EI-Safe-MADDPG  still has the best  emergent incident response completion rate for inbound and outbound operations, at 92.9\%. The emergent incident response completion rate of C-Safe-MADDPG   is second-highest, with a maximum of 59.9\%. The emergent incident response completion rate of Safe-MADDPG   is still the worst, at around 50.3\%. From Fig.  \ref{fig:6.2}(d), it can be seen that the emergent incident response loss rate of   EI-Safe-MADDPG   for inbound and outbound operations   is the lowest among the three algorithms, which can reach 6\%. The emergent incident response loss rate of C-Safe-MADDPG   is second-lowest, with a minimum of 45.3\%. The emergent incident response loss rate of Safe-MADDPG   is the worst,   at around 53.1\%.

\begin{figure}[htp]
    \centering
    \includegraphics[width=9cm]{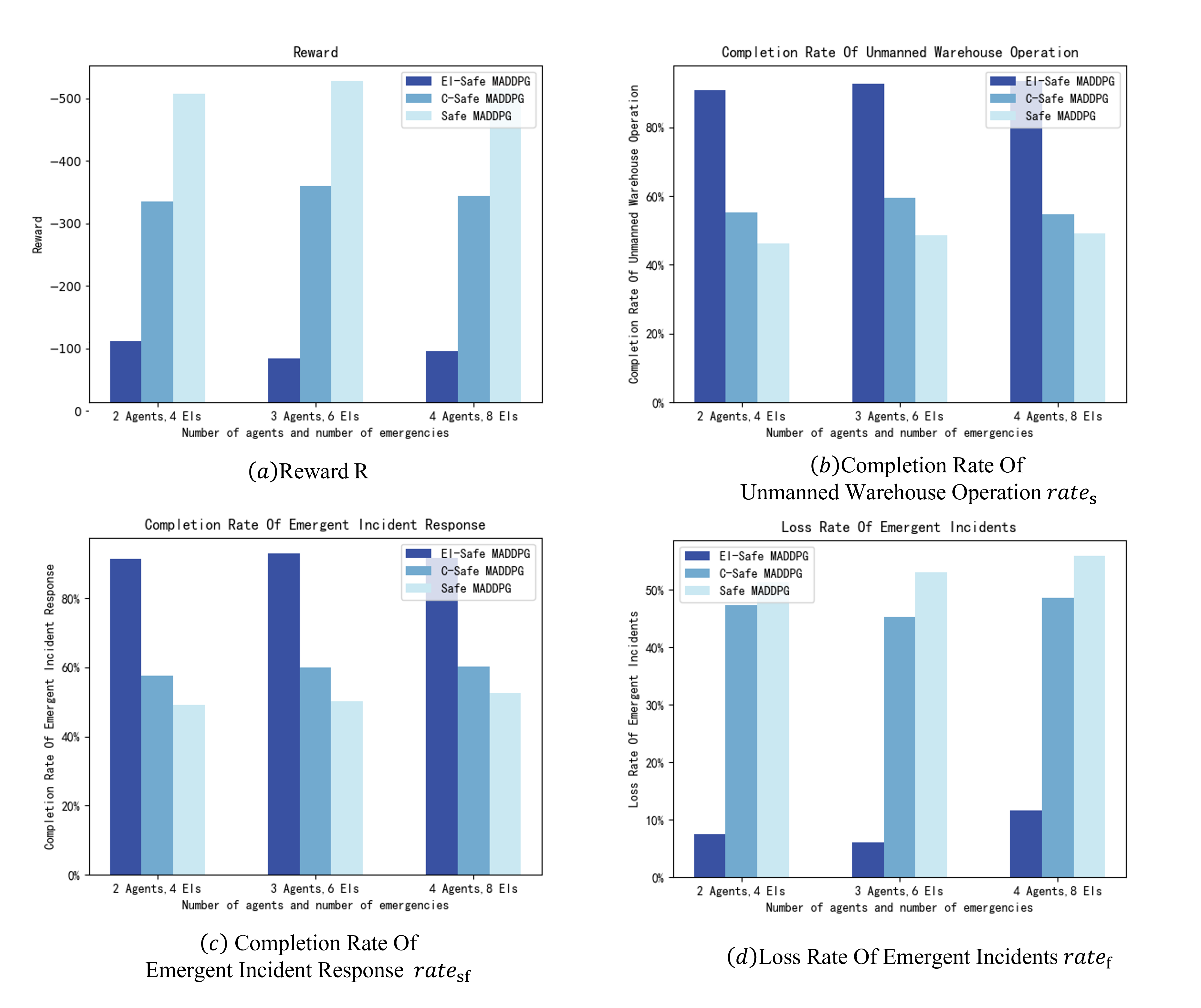}
    \caption{Experimental results}
    \label{fig:6.3}
\end{figure}

We also conducted experiments on  layout $A$ with 2 agents for emergent incident response and 4 emergent incidents, as well as   with 4 agents for emergent incident response and 8 emergent incidents. Fig. \ref{fig:6.3} compares the experimental results   of EI-Safe-MADDPG, C-Safe-MADDPG, and Safe-MADDPG  under the three parameter settings. Fig. \ref{fig:6.3}(a) shows the average reward function $\overline{r}$ under the three parameter settings. Fig. \ref{fig:6.3}(b) shows the average completion rate  $\overline{rate_{s}}$ of inbound and outbound operations   under the three parameter settings,   Fig. \ref{fig:6.3}(c) shows the average emergent incident response completion rate $\overline{rate_{sf}}$, and Fig. \ref{fig:6.3}(d) shows the average emergent incident response loss rate $\overline{rate_{f}}$.
\begin{table*}[htbp]
\caption{Layout scene inbound/outbound operation experiment results}
\begin{center}
\begin{tabular}{|c|c|c|c|c|c|}
\hline
\textbf{Algorithm} & \textbf{Number of agents and emergencies}& \textbf{$\overline{rate_{s}}$}& \textbf{$\overline{rate_{f}}$}& \textbf{$\overline{rate_{sf}}$}& \textbf{$\overline{r}$}\\
\hline
\multirow{3}{*}{EI-Safe-MADDPG}
& 2, 4 & 90.9\% & 7.5\% & 91.3\% & -102.26 \\
\cline{2-6} 
& 3, 6 & 92.7\% & 6\% & \textbf{\red{92.9\%}} & \textbf{\red{-72.4}} \\ 
\cline{2-6} 
& 4, 8 & \textbf{\red{93.4\%}} & \textbf{\red{5.7\%}} & 91.6\% & -85.354 \\
\hline
\multirow{3}{*}{C-Safe-MADDPG}
& 2, 4 & 55.3\% & 47.4\% & 57.6\% & -334.03 \\
\cline{2-6} 
& 3, 6 & 59.5\% & 45.3\% & 59.9\% & -360.5 \\ 
\cline{2-6} 
& 4, 8 & 54.7\% & 48.6\%	& 60.2\% & -342.93 \\
\hline
\multirow{3}{*}{Safe-MADDPG}
& 2, 4 & 46.3\% & 51.3\% & 49.1\% & -513.96 \\
\cline{2-6} 
& 3, 6 & 48.5\% & 53.1\% & 50.3\% & -534.2 \\ 
\cline{2-6} 
& 4, 8 & 49.2\% & 55.9\%	& 52.6\% & -522.75 \\
\hline
\end{tabular}
\label{tab2}
\end{center}
\end{table*}
In Fig. \ref{fig:6.3}, each bar chart represents  $\overline{rate_{s}}$,  $\overline{rate_{f}}$,  $\overline{rate_{sf}}$, and average reward function $\overline{r}$ when the training episodes are set to 20,000 for each scenario. According to the experimental results,  $\overline{rate_{s}}$,  $\overline{rate_{f}}$,  $\overline{rate_{sf}}$, and  $\overline{r}$ of the nine experiments using  EI-Safe-MADDPG  are the best. Table \ref{tab2} summarizes the average values of  evaluation indicators of different algorithms and experimental parameter settings, and highlights the best   results  in red boldface.

From Table \ref{tab2}, it can be seen that among all algorithms and experimental settings,   $\overline{rate_{s}}$ is   highest, reaching 93.4\%, when the number of emergent incident response agents is 4, the number of emergent incidents is 8, and the algorithm is EI-Safe-MADDPG; and $\overline{rate_{f}}$ is   lowest, at 5.7\%, when the number of agents is 4, the number of emergent incidents is 8, and the algorithm is EI-Safe-MADDPG.   $\overline{rate_{sf}}$ is   highest, reaching 92.9\%, when the number of agents is 3, the number of emergent incidents is 6, and the algorithm is EI-Safe-MADDPG. The highest average reward function $\overline{r}$  is -72.4 when the number of agents is 3, the number of emergent incidents is 6, and the algorithm is EI-Safe-MADDPG. In summary,   the proposed EI-Safe-MADDPG algorithm  provides the best experimental results under three   experimental configurations in unmanned warehouse layout $A$, with the highest average completion rate $\overline{rate_{s}}$,   lowest average emergent incident response loss rate $\overline{rate_{f}}$,   highest average emergent incident response completion rate $\overline{rate_{sf}}$, and   highest average reward function $\overline{r}$.
\begin{table*}[htbp]
\caption{Experimental results of different layouts and experimental settings of unmanned warehouse}
\begin{center}
\begin{tabular}{|c|c|c|c|c|c|c|}
\hline
\textbf{Algorithm} & Layout type & \textbf{Number of agents and emergencies}& \textbf{$\overline{rate_{s}}$}& \textbf{$\overline{rate_{f}}$}& \textbf{$\overline{rate_{sf}}$}& \textbf{$\overline{r}$}\\
\hline
\multirow{6}{*}{EI-Safe-MADDPG}
& \multirow{3}{*}{B}
& 2, 4 & \textbf{\red{93.1\%}} & \textbf{\red{8.4\%}} & \textbf{\red{89.2\%}} & -114.281 \\
\cline{3-7} 
&& 3, 6 & 89.9\% & 9.8\% & 88.9\% & \textbf{\red{-71.625}} \\ 
\cline{3-7} 
&& 4, 8 & 92.5\% & 10.6\% & 88.3\% & -82.294 \\
\cline{2-7} 

& \multirow{3}{*}{C}
& 2, 4 & 92.7\% & \textbf{\red{7.9\%}} & 91.1\% & -126.351 \\
\cline{3-7} 
&& 3, 6 & 90.6\% & 8.5\% & 90.5\% & \textbf{\red{-70.633}} \\ 
\cline{3-7} 
&& 4, 8 & \textbf{\red{93.4\%}} & 8.8\% & \textbf{\red{91.7\%}} & -83.192 \\
\hline

\multirow{6}{*}{C-Safe-MADDPG}
& \multirow{3}{*}{B}
& 2, 4 & 52.5\% & 44.4\% & 55.4\% & -345.45 \\
\cline{3-7} 
&& 3, 6 & 50.7\% & 46.2\% & 59.6\% & -332.67 \\ 
\cline{3-7} 
&& 4, 8 & 53.5\% & 45.9\% & 57.8\% & -348.36 \\
\cline{2-7} 

& \multirow{3}{*}{C}
& 2, 4 & 58.2\% & 40.7\% & 55.7\% & -356.43 \\
\cline{3-7} 
&& 3, 6 & 55.3\% & 43.64\% & 53.2\% & -363.61 \\ 
\cline{3-7} 
&& 4, 8 & 57.5\% & 44.1\% & 57.1\% & -359.28 \\
\hline

\multirow{6}{*}{Safe-MADDPG} 
& \multirow{3}{*}{B}
& 2, 4 & 48.5\% & 56.1\% & 51.7\% & -531.72 \\
\cline{3-7} 
&& 3, 6 & 40.4\% & 55.2\% & 56.7\% & -550.16 \\ 
\cline{3-7} 
&& 4, 8 & 41.7\% & 59.7\% & 54.3\% & -570.83 \\
\cline{2-7} 

& \multirow{3}{*}{C}
& 2, 4 & 42.1\% & 54.3\% & 50.7\% & -502.17 \\
\cline{3-7} 
&& 3, 6 & 49.8\% & 53.9\% & 57.9\% & -567.93 \\ 
\cline{3-7} 
&& 4, 8 & 40.9\% & 51.5\% & 56.4\% & -524.61 \\
\hline

\end{tabular}
\label{tab3}
\end{center}
\end{table*}

We also conducted experiments on   layouts $B$ and   $C$, with 2 emergent incident response agents and 4 emergent incidents, 3 emergent incident response agents and 6 emergent incidents, and 4 emergent incident response agents and 8 emergent incidents for each layout, with results of the four evaluation indicators under different experimental parameter settings for each layout as shown in Table \ref{tab3}.

Among all unmanned warehouse layout types and experimental settings, from Table \ref{tab2}, it can be seen that   EI-Safe-MADDPG   provides the best experimental results in     layout $A$.

From Table \ref{tab3}, it can be seen that in   layout $B$,   $\overline{rate_{s}}$ is   highest, reaching 93.1\%, when the number of emergent incident response agents is 2, the number of emergent incidents is 4, and the algorithm is EI-Safe-MADDPG.   $\overline{rate_{f}}$ is   lowest, at 8.4\%, when the number of agents is 2, the number of emergent incidents is 4, and the algorithm is EI-Safe-MADDPG.   $\overline{rate_{sf}}$ is   highest, reaching 89.2\%, when the number of agents is 2, the number of incidents is 4, and the algorithm is EI-Safe-MADDPG.   $\overline{r}$  is highest, at -71.625, when the number of agents is 3, the number of emergent incidents is 6, and the algorithm is EI-Safe-MADDPG. In summary, according to the experimental results, the proposed EI-Safe-MADDPG   provides the best experimental results under three   experimental configurations in   warehouse layout $B$, with the highest   $\overline{rate_{s}}$,   lowest  $\overline{rate_{f}}$,   highest   $\overline{rate_{sf}}$, and   highest   $\overline{r}$.

In   warehouse layout $C$,   $\overline{rate_{s}}$ is   highest, reaching 93.4\%, when the number of emergent incident response agents is 4, the number of emergent incidents is 8, and the algorithm is EI-Safe-MADDPG.   $\overline{rate_{f}}$ is   lowest, at 7.9\%, when the number of agents is 2, the number of emergent incidents is 4, and the algorithm is EI-Safe-MADDPG.   $\overline{rate_{sf}}$ is   highest, reaching 91.7\%,  when the number of agents is 4, the number of incidents is 8, and the algorithm is EI-Safe-MADDPG. The highest   $\overline{r}$ is -70.633, when the number of agents is 3, the number of emergent incidents is 6, and the algorithm is EI-Safe-MADDPG. In summary, according to the experimental results, the proposed EI-Safe-MADDPG algorithm   provides the best experimental results under three   experimental configurations in unmanned warehouse layout $C$, with the highest  $\overline{rate_{s}}$,  lowest  $\overline{rate_{f}}$,  highest  $\overline{rate_{sf}}$, and  highest  $\overline{r}$.
Under different unmanned warehouse layout types and experimental settings, the proposed EI-Safe-MADDPG algorithm, based on Safe-MADDPG for emergent incident response coordination optimization, provided the best  results, with the highest   $\overline{rate_{s}}$,   lowest   $\overline{rate_{f}}$,   highest   $\overline{rate_{sf}}$, and   highest   $\overline{r}$.

\section{Conclusion}
In this paper, we modeled the emergency decision problem and proposed an EI-Safe-MADDPG algorithm combining a constraint-extraction algorithm to extract security constraints and obtain better security performance, in addition to Safe-MADDPGfor discussing the emergency decision problem of a multi-agent rescue system.  Simulation environments of different layouts were evaluated, and the performance of the decision method in different environments was   experimentally compared, which confirmed the effectiveness of the EI-Safe-MADDPG algorithm.


\end{CJK}
\bibliography{ref}
\end{document}